%% file: arxiv-main.tex
\documentclass[sigconf,authorversion]{acmart}

\usepackage{algorithm}
\usepackage{algorithmic}

\usepackage[absolute]{textpos}

\AtBeginDocument{%
  \providecommand\BibTeX{{%
    \normalfont B\kern-0.5em{\scshape i\kern-0.25em b}\kern-0.8em\TeX}}}

\usepackage{newfloat}
\usepackage{listings}

\usepackage{subcaption} 
\usepackage{graphicx}
\usepackage{listings}

\setcopyright{none}
\renewcommand\footnotetextcopyrightpermission[1]{}
\makeindex

\begin{document}

\iffalse
\begin{textblock}{15}(0.6,1)
\noindent\small {\bf This article is currently in submission and under review. Please do not distribute.}
%\noindent\small This article is submitted to the Thirty-Eighth AAAI Conference on Artificial Intelligence (AAAI-24) and is currently under review. Please do not distribute.
\end{textblock}
\fi

\title{Measuring Distributional Shifts in Text: The Advantage of Language Model-Based Embeddings}
\author{Gyandev Gupta\textsuperscript{\rm 1}\textsuperscript{\rm *}, Bashir Rastegarpanah\textsuperscript{\rm 2}\textsuperscript{\rm *}, Amalendu Iyer\textsuperscript{\rm 2}, Joshua Rubin\textsuperscript{\rm 2}, Krishnaram Kenthapadi\textsuperscript{\rm 2}\\
\textsuperscript{\rm 1}Indian Institute of Technology - Bombay, \textsuperscript{\rm 2}Fiddler AI \ \  \textsuperscript{\rm *}Equal Contribution
}
%\iffalse
\affiliation{
    %\textsuperscript{\rm 1} Indian Institute of Technology - Bombay\\
    %\textsuperscript{\rm 2} Fiddler AI\\
    %gyandevgupta.iitb23@gmail.com, \{bashir, amal, josh, krishnaram\}@fiddler.ai
    gyandevgupta.iitb23@gmail.com, bashir@bu.edu, amal@fiddler.ai, josh@fiddler.ai, krishnaram@fiddler.ai
}
%\fi

\renewcommand{\shortauthors}{Gupta, et al.}

\begin{abstract}
\input{0-abstract.tex}
\end{abstract}

\maketitle
\pagestyle{plain}

\section{Introduction}\label{sec:intro}
\input{1-introduction.tex}

\section{Related Work}\label{lifecycle}
\input{2-background-and-related-work.tex}

% \section{Design and Architecture}\label{design}
% \input{3-modelmonitor-design-and-architecture.tex}

\section{Data Drift Measurement Methodology}\label{sec:methodology}
\input{3-vector-monitoring-method.tex}

\section{Empirical Evaluations}\label{sec:quantitative}
\input{4-evalutations.tex}

\section{Deployment Case Study}\label{case-study}
\input{5-case-study.tex}

%\section{Deployment Insights and Lessons Learned in Practice}\label{use-cases}
\section{Deployment Insights and Lessons Learned}\label{use-cases}
\input{6-use-cases.tex}

%\section{Related Work}\label{sec:related}
%\input{7-related-work.tex}

%\section{Summary and Concluding Remarks}\label{sec:conclusion}
\section{Conclusion}\label{sec:conclusion}
\input{8-conclusion.tex}

%% dropping to preserve anonymity during submission -- we can add to the final version

\iffalse
\section{Acknowledgments}
The authors would like to thank other members of the Fiddler AI team for their collaboration
during the development and deployment of the approach for measuring distributional shifts
in unstructured and natural language data. 
In particular, we would like to thank 
Gabriel Atkin
Lior Berry,
Danny Brock,
Nilesh Dalvi,
Krishna Gade,
Lea Genuit,
Barun Halder,
Karen He,
Morris Hsu,
Amit Paka,
Thuy Pham,
Mary Reagan,
Murtuza N. Shergadwala,
and
Anu Vatsa, for insightful discussions and feedback.
Furthermore, we would like to thank all Fiddler AI customers whose monitoring
use cases helped us design the solutions presented in this paper.
\fi

\bibliographystyle{ACM-Reference-Format}
\bibliography{bibfile.bib}

%\newpage
\clearpage

%\appendix

%\input{appendixfile.tex}

\end{document}

%% file: 0-abstract.tex
An essential part of monitoring machine learning models in production is measuring input and output data drift. In this paper, we present a system for 
measuring distributional shifts in natural language data and highlight and investigate the potential advantage of using large language models (LLMs) for this problem. Recent advancements in LLMs and their successful adoption in different domains indicate their effectiveness in capturing semantic relationships for solving various natural language processing problems. The power of LLMs comes largely from the encodings (embeddings) generated in the hidden layers of the corresponding neural network. First we propose a clustering-based algorithm for measuring distributional shifts in text data by exploiting such embeddings. Then we study the effectiveness of our approach when applied to text embeddings generated by both LLMs and classical embedding algorithms. Our experiments show that general-purpose LLM-based embeddings provide a high sensitivity to data drift compared to other embedding methods. We propose drift sensitivity as an important evaluation metric to consider when comparing language models. Finally, we present insights and lessons learned from deploying our framework as part of the Fiddler ML Monitoring platform over a period of 18 months.

%% file: 1-introduction.tex
With the increasing deployment of AI systems in high-stakes settings such as healthcare, autonomous systems, lending, hiring, and education, we need to make sure that the underlying machine learning (ML) models are not only accurate but also reliable, robust, and deployed in a trustworthy manner. While ML models are typically validated and stress-tested before deployment, they may not be sufficiently scrutinized after deployment. A common assumption when deploying ML models is that the underlying data distribution observed during deployment remains unchanged compared to that of the training data. The violation of this assumption often results in unexpected model behaviors and can potentially cause subsequent changes, such as degradation in model performance or unacceptable model outputs. Hence, early detection of distributional shifts in data observed during deployment is
an essential part of monitoring deployed ML systems \cite{breck2019data, kurshan2020towards, makinen2021needs, sculley2015hidden, shankar2022operationalizing, shergadwala2022human}. In many real-world settings, it may not be feasible to obtain immediate user feedback signals or human judgments, or incorporate other mechanisms for continuous assessment of the quality of the ML model. In such scenarios, detecting distributional shifts could serve as an early indicator of the model's performance degradation or unexpected behavior and help protect against potential catastrophic failures, e.g., by gracefully deferring to human experts in the case of human-AI hybrid systems.

Natural language processing (NLP) models are increasingly being used in ML pipelines, and the advent of new technologies, such as large language models (LLMs), has greatly extended the adaption of NLP solutions in different domains. Consequently, the problem of distributional shift (``data drift'') discussed above must be addressed for NLP data to avoid performance degradation after deployment \cite{wang2022measure}. Examples of data drift in NLP data include the emergence of a new topic in customer chats, the emergence of spam or phishing email messages that follow a different distribution compared to that used for training the detection models, and differences in the characteristics of prompts issued to an LLM application during deployment compared to the characteristics of prompts used during testing and validation.

In the case of structured data consisting of categorical and numerical features,
a common approach to monitor data drift 
is to estimate the univariate distribution of input features
(e.g., by using a histogram binning)
for both a baseline dataset and the data at production time,
and then quantify drift using distributional
distance metrics such as the Jensen-Shannon divergence (JSD).
However, in the case of unstructured data like text,
estimating the data distribution using histogram binning
is not a trivial task.

Modern NLP pipelines process text inputs in steps.
Text is typically converted to tokens,
which are mapped into a continuous (high-dimensional) embedding space to get 
a numerical vector representation, which ML models can consume.
Distributional shifts in text data can be
measured in this embedding space.
However, estimating the data distribution using
a binning procedure becomes challenging when the
data is in a high-dimensional space 
%like text embeddings 
for the following reasons.
The number of bins increases exponentially with the number of dimensions, and the appropriate binning resolution cannot be selected a priori.
%That is because the number of bins can easily explode as the number of dimensions increases, and the appropriate binning resolution cannot be selected a priori.

%The first key contribution of this paper is proposing 
We propose 
a novel method for measuring distributional shifts in
embedding spaces.
We use a data-driven approach to detect high-density
regions in the embedding space of the baseline 
%(reference) 
data,
and track how the relative density of such regions
changes over time.
In particular, we apply k-means clustering
to partition the embedding space into disjoint regions of high-density
in the baseline data.
The cluster centroids are then used to design a binning strategy
which allows us to calculate a drift value for subsequent observations.

As in many other NLP problems, having access to high-quality text embeddings
is crucial for measuring data drift as well.
The more an embedding model is capable of capturing semantic relationships,
the higher the sensitivity of clustering-based drift monitoring.
Recent developments in LLMs have shown that the embeddings that are generated internally by such models are
%very powerful in
capable of capturing semantic relationships.
In fact, some LLMs are specifically trained to provide general-purpose text embeddings.

%The next key contribution of this paper is 
We perform
an empirical study for comparing the 
effectiveness of different embedding models in measuring distributional shifts in text.
We use three real-world text datasets and apply our proposed clustering-based algorithm
to measure both the existing drift in the real data and the synthetic drift that we 
introduce by modifying the distribution of text data points.
We compare multiple embedding models (both classical and LLM-based models) and
measure their sensitivity to distributional shifts at different levels of data drift.
Our experiments show that LLM-based embeddings in general outperform classical
embeddings when sensitivity to data drift is considered. 
We also highlight insights and lessons learned from deploying our system as part of the Fiddler ML Monitoring platform \cite{fiddler_monitor} for 18 months.
%We also highlight insights and lessons learned from deploying these approaches as part of the Fiddler ML monitoring platform a model monitoring platform.

In summary, the key contributions of this paper are:
\begin{itemize}
  \item Proposing a novel clustering-based method and system for measuring data drift in embedding spaces.
  \item Highlighting the application of LLM-based embeddings for data drift monitoring.
  \item Introducing \emph{sensitivity to drift} as an evaluation metric for comparing embeddings models.
  \item An empirical study of the effectiveness of different embedding models in detecting data drift.
  \item Presenting insights and lessons learned from deploying our system in practice.
\end{itemize}

%% file: 2-background-and-related-work.tex
{\noindent \bf Distributional Shift Detection for Categorical and Numerical Data} There is extensive work on detecting distributional shifts in categorical and numerical data (refer \citet{breck2019data, cormode2021relative, gama2014survey, karnin2016optimal, rabanser2019failing, tsymbal2004problem, webb2016characterizing, vzliobaite2016overview} for an overview). For practical ML applications, there is often a need to verify the validity of model inputs, which can be done by checking if the value is within a specified range and performing other user-defined tests \cite{schelter2018automating}. Statistical hypothesis testing and confidence interval-based approaches are often used to detect changes in the features or model outputs. While simpler tests such as Student's t-test and Kolmogorov–Smirnov test are employed for univariate or low dimensional data \cite{murphy2012machine, wasserman2004all}, advanced tests such as Maximum Mean Discrepancy are useful for higher dimensional data \cite{gretton2012kernel}. As hypothesis testing approaches require fine-tuning, confidence interval-based approaches are often preferred in practice  \cite{efron1994introduction, nigenda2022amazon}. In general, the above change detection methods require density estimation, which is commonly obtained using histograms \cite{silverman2018density}. For multivariate data, the number of histogram bins grows exponentially with the dimensionality of data, and hence tree-based \cite{boracchi2017uniform, boracchi2018quanttree} and clustering-based \cite{liu2020concept} data partitioning approaches have been proposed to scale well for high-dimensional data. While the clustering-based approach proposed in \citet{liu2020concept} has similarities to our approach, this work focuses exclusively on multi-variate categorical and numerical data whereas our main focus is on measuring distributional shifts in text. Finally, while our approach is designed to be model-agnostic, there is also work on making use of model internals to detect drifts and take corrective actions \cite{NEURIPS2020_219e0524,lipton2018detecting,reddi2015doubly,wu2019domain}.\\

{\noindent \bf Robustness in NLP Models} In light of the recent advances in NLP models (including LLMs) and the associated practical applications, there is a rich literature on techniques for measuring and improving robustness in NLP models (see \cite{wang2022measure} and references therein). This line of work is motivated by the fact that NLP models (including LLMs) are often brittle to out-of-domain data, adversarial attacks, or small perturbations to the input. While this work largely pertains to measuring robustness prior to deploying NLP models, our work focuses on the related but complementary problem of measuring distributional shifts in text once the models are deployed.\\

% TODO: Nice to have: include relevant references from \cite{wang2022measure} and from ICML 2023 tutorial

% {\noindent \bf LLM background and getting embeddings from LLMs.  LLMs trained for providing general purpose emebddings}
% TODO: Nice-to-have section: to be revisited and expanded, time-permitting and to be dropped otherwise.
% \\

%Detecting and Correcting for Label Shift with Black Box Predictors \cite{lipton2018detecting}\\
%A Novel Non-parametric Two-Sample Test on Imprecise Observations\cite{liu2020novel}

{\noindent \bf Tools for Monitoring Deployed ML Models} Several open source and commercial tools provide users with the ability to monitor predictions of deployed ML models, e.g., Amazon SageMaker Model Monitor \cite{nigenda2022amazon}, Arize Monitoring \cite{arize_monitor}, Deequ \cite{schelter2018automating}, Evidently \cite{evidently}, Fiddler's Explainable Monitoring \cite{fiddler_monitor}, Google Vertex AI Model Monitoring \cite{google_drift}, IBM Watson OpenScale \cite{ibm_monitor}, Microsoft Azure MLOps \cite{azure_monitor}, TruEra Monitoring \cite{truera_monitor}, and Uber's Michelangelo platform \cite{uber_michelangelo}. 
%Our work on measuring distributional shifts in text is broadly applicable in the context of such tools.

%% file: 3-vector-monitoring-method.tex
% Unstructured data types including image and text data are usually represented as
% multi-dimensional vectors when used as inputs to ML models.
% The vector representation is generally achieved through a transformation step
% that maps original data points to an embedding space that satisfies certain
% desired properties
% (e.g., text embeddings are designed to preserve semantic relationships).
% Consequently, measuring distributional shifts in unstructured data boils down
% to the capability of tracking distributional shifts in multi-dimensional vector spaces.
Unstructured data such as images and text can be represented as semantically sensitive vectors
by a variety of means.  In natural language, these vectors can be simple term frequencies or summed word-level embeddings from off-the-shelf libraries.  Increasingly practitioners use internal, context-sensitive, learned representations computed by deep learning models for this task as they are maximally sensitive to application specific semantics.  Taking advantage of this semantic sensitivity, we treat the problem of unstructured data drift as tracking distributional shift in multi-dimensional vector distributions.

Histogram-based methods are commonly used for measuring distributional shifts where 
distributional distance metrics such as the Jensen-Shannon divergence (JSD)
are applied to quantify data drift.
In this approach, one needs to first find a histogram approximation of the two distributions
at hand. In the case of univariate tabular data (i.e., one-dimensional distributions),
generating these histograms is fairly straightforward and is achieved via a binning procedure
where data points are assigned to histogram bins defined as a particular interval of the variable range.

However,
generalizing this procedure to higher dimensional distributions requires identifying an efficient multivariate binning strategy.
% working with vector representations of unstructured data,
% the above binning procedure is not practical since finding an efficient binning procedure
% for multi-dimensional vector distributions is not a trivial task.
Grid-based space partitioning algorithms are impractical as the number of bins increases exponentially with the number of dimensions. 
%Grid-based space partitioning algorithms are impractical as they can leave many empty bins and explode as the number of dimensions increases.
Tree-based approaches are inefficient as, unlike for structured data, structure in semantic vector distributions is rarely axis-aligned.
% Binning methods such as grid-based or tree-based space partitioning algorithms 
% are not always practical since the appropriate binning resolution cannot be set a priori.
% Furthermore, the number of bins can easily explode as the number of dimensions increases. Therefore, the main challenge in monitoring vector data is
% finding an efficient and data-driven binning procedure for multi-dimensional distributions. 

\subsection{Clustering-based Vector Monitoring}
The core idea behind our vector monitoring method is a novel binning procedure
in which instead of using fixed interval bins,
bins are defined as regions of high-density in the data space.
The density-based bins are automatically detected using standard clustering
algorithms such as k-means.
Once we achieve the histogram bins for both baseline and production data,
we can apply any of the distributional distance metrics used for measuring the
discrepancy between two histograms.
In the following we provide an step by step introduction to our vector monitoring
algorithm using an illustrative example.

\begin{figure}[h]
\centering
  \includegraphics[width=0.7\columnwidth]{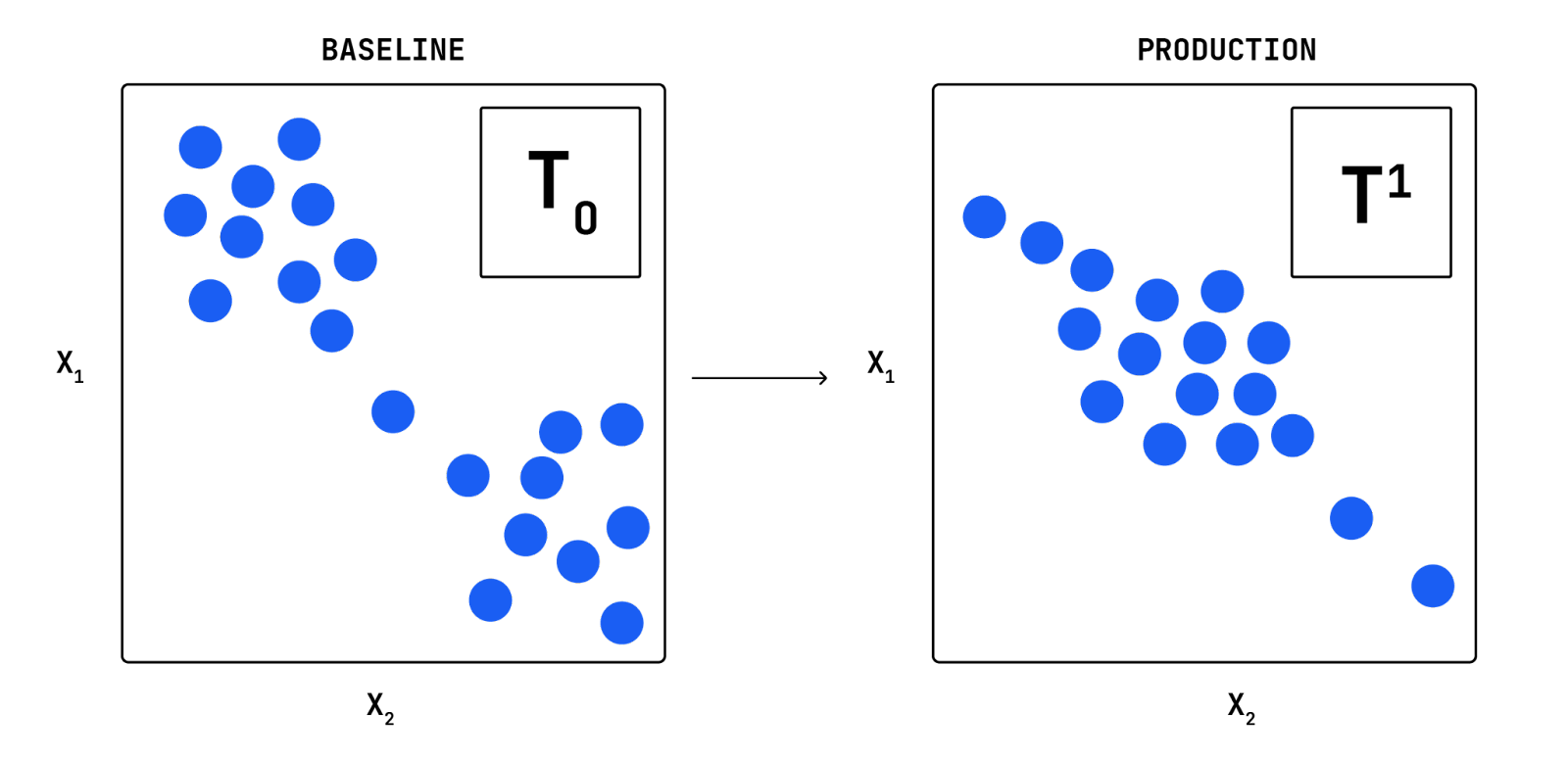}
  \caption{A data drift scenario in two dimensions.}
  \label{fig:method-data}
\end{figure}

Consider the example of a multi-dimensional data drift scenario presented
in Figure \ref{fig:method-data} where,
for the sake of simplicity,
the vector data points are 2-dimensional.
Comparing the baseline data (left plot) with the example production data (right plot), we see a shift in the data distribution where more data points are located around the center of the plot. Note that in practice the vector dimensions are usually much larger than two and such a visual diagnosis is impossible.
We would like to have an automatic procedure that precisely quantifies the amount of data drift in a scenario like this.

The first step of our clustering-based drift detection algorithm is to
detect regions of high density (data clusters) in the baseline data.
We achieve this by taking all the baseline vectors and partitioning them into a
fixed number of well populated clusters using the k-means clustering algorithm. 

\begin{figure}[h]
\centering
  \includegraphics[width=\columnwidth]{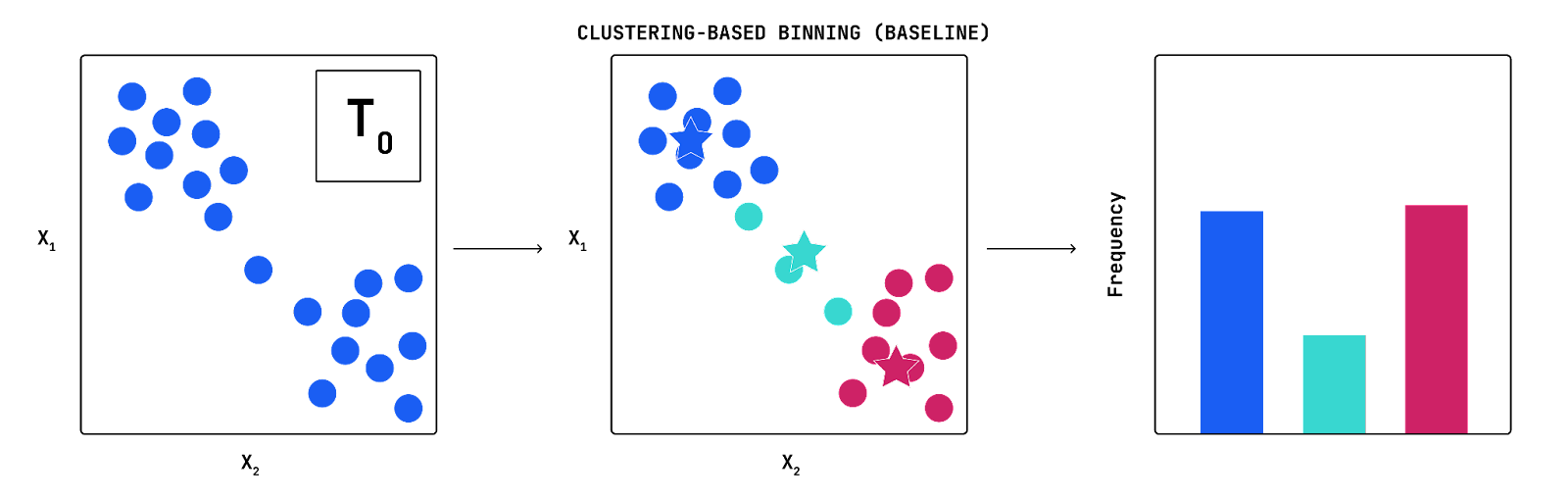}
  \caption{Applying k-means algorithm on baseline data.}
  \label{fig:method-baseline}
\end{figure}

Figure \ref{fig:method-baseline} shows the output of the clustering step (k=3)
applied to our illustrative example where data points are colored by their
cluster assignments.
After baseline data are partitioned into clusters, the relative frequency of
data points in each cluster (i.e., the relative cluster size)
is assigned to the corresponding histogram bin.
As a result, we obtain a 1-dimensional binned histogram of baseline data.

As our goal is to monitor for shifts in the data distribution, we track how the relative data density changes over time in different partitions (clusters) of the space.
%As we mentioned earlier, our goal is to monitor for shifts in the data distribution. We achieve this via tracking how the relative data density changes over time in different partitions (clusters) of the space.
The number of clusters can be interpreted as the resolution by which the drift monitoring will be performed; the higher the number of clusters, the higher the sensitivity to data drift.

\begin{figure}[h]
\centering
  \includegraphics[width=\columnwidth]{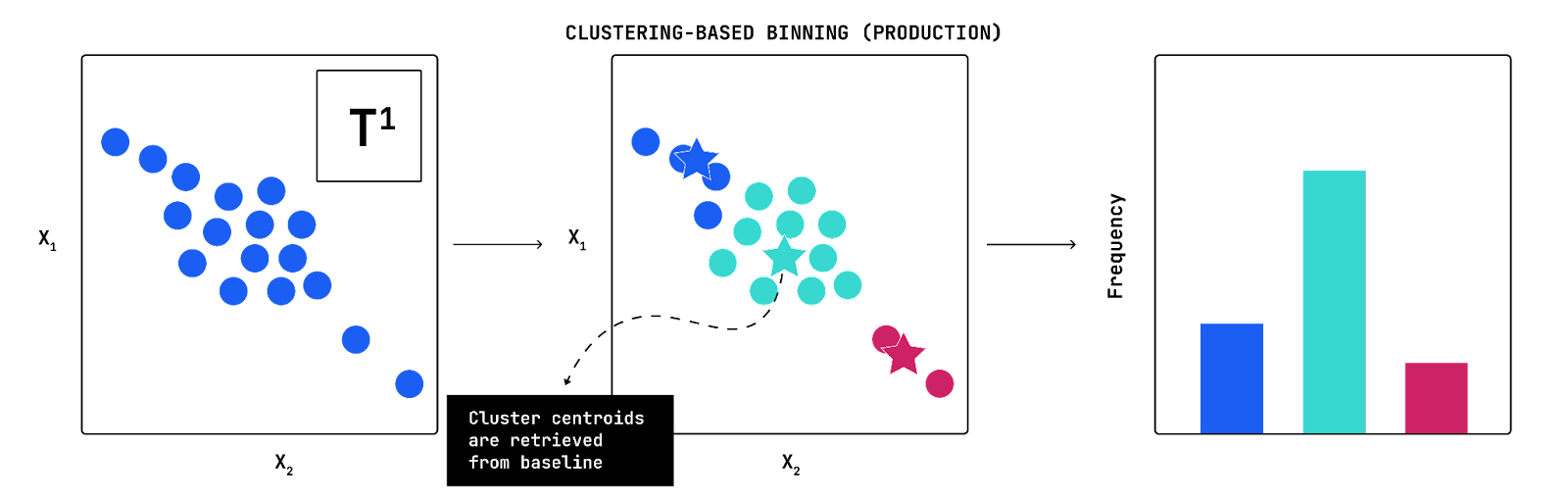}
  \caption{Assigning production data to baseline clusters.}
  \label{fig:method-production}
\end{figure}

After running k-means algorithm on the baseline data with a given number of clusters $k$, we obtain $k$ cluster centroids. We use these cluster centroids in order to generate the binned histogram of the production data. In particular, fixing the cluster centroids detected from the baseline data, we assign each incoming data point to the bin whose cluster centroid has the smallest distance to the data point.
Applying this procedure to the example production data 
%previously shown in 
from
Figure \ref{fig:method-data},
and normalizing the bins,
we obtain the cluster frequency histogram for the production data 
%that is presented in 
(Figure \ref{fig:method-production}).

\begin{figure}[h]
\centering
  \includegraphics[width=0.7\columnwidth]{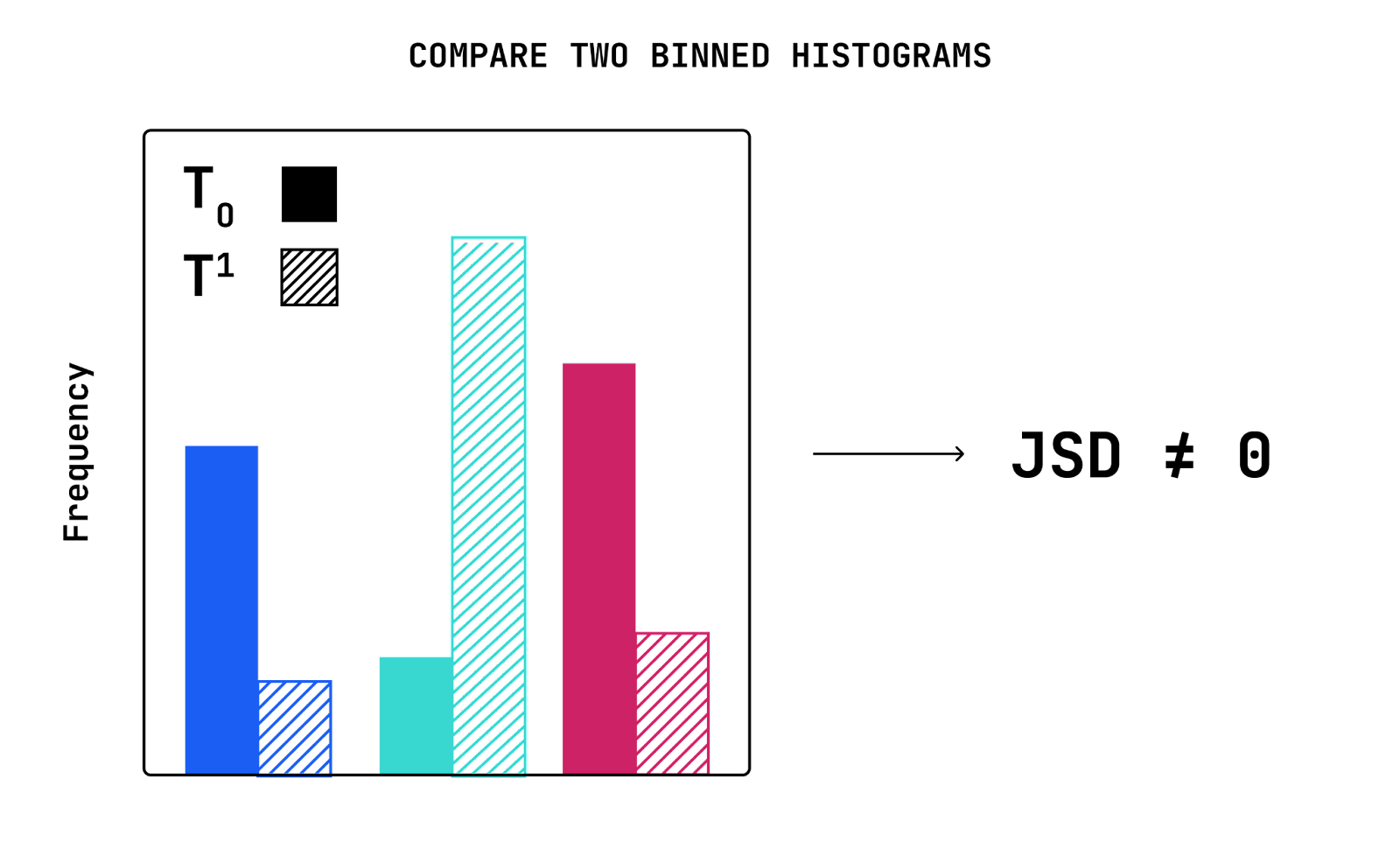}
  \caption{Quantify data drift using distribution distance metrics.}
  \label{fig:method-compare}
\end{figure}

Finally, we can use a conventional distance measure
like JSD between baseline and production histograms to get
a drift value (Figure \ref{fig:method-compare}).
This drift value helps identify any changes in the relative density
of cluster partitions over time.

\begin{algorithm}[t]
\caption{\texttt{InitializeClusters}}
\label{alg:init}
\begin{flushleft}
\textbf{Input}: baseline dataset $D_{base}$, number of clusters $k$\\
%\textbf{Parameter}: Optional list of parameters\\
\textbf{Output}: centroids $C= [c_1, \dots, c_k]$,
normalized frequencies $F = [f_1, \dots, f_k]$
\end{flushleft}
\begin{algorithmic}[1] %[1] enables line numbers
\STATE Apply \texttt{k-means($n_{clusters}=k$)} on $D_{base}$
\STATE Let $C :=$ cluster centroids found by \texttt{k-means}
\STATE Initialize $F$ with a size $k$ vector of 0's. 
\FOR{$d$ in $D_{base}$}
\STATE find the centroid $c_i \in C$ that minimizes $||c_i - d||_2$  
\STATE Let $F[i] := F[i]+1$
\ENDFOR
\STATE $F := \frac{F}{|D_{base}|}$
\STATE \textbf{return} $C$, $F$
\end{algorithmic}
\end{algorithm}

\begin{algorithm}[t]
\caption{\texttt{ComputeDrift}}
\label{alg:jsd}
\begin{flushleft}
\textbf{Input}: baseline dataset $D_{base}$,
production dataset $D_{prod}$,
number of clusters $k$\\
%\textbf{Parameter}: Optional list of parameters\\
\textbf{Output}: drift value $v$
\end{flushleft}
\begin{algorithmic}[1] %[1] enables line numbers
\STATE $C, F_{base} :=$ \texttt{InitializeClusters($D_{base}, k$)}
\STATE Initialize $F_{prod}$ with a size $k$ vector of 0's. 
\FOR{$d$ in $D_{prod}$}
\STATE find the centroid $c_i \in C$ that minimizes $||c_i - d||_2$  
\STATE Let $F_{prod}[i] := F_{prod}[i]+1$
\ENDFOR
\STATE $F_{prod} := \frac{F_{prod}}{|D_{prod}|}$
\STATE \textbf{return} \texttt{Jensen-Shannon-Divergence$(F_{base},F_{prod}$)} 
\end{algorithmic}
\end{algorithm}

Algorithms \ref{alg:init} and \ref{alg:jsd} present the above procedure
for computing data drift for a given baseline dataset $D_{base}$ and
a dataset $D_{prod}$ of observations at deployment for which we want 
to measure the distributional shift.

\subsection{Choosing the Number of Clusters}
Running the k-means algorithm requires specifying the number of 
clusters. Since we use clustering as a method to partition 
the embedding space for measuring data drift, the number of clusters
in our algorithm can be seen as a tuning parameter that specifies
the resolution of our measurement. That is, the higher the number 
of clusters, the higher the sensitivity of our measurement method to 
shifts in the data distribution.

From a practical perspective, the number of clusters can be increased
until there are enough data points in each cluster such that 
there is sufficient (statistical) evidence for a measured drift value. Therefore, a binary search can be used to find the 
largest number of clusters that ensures sufficient evidence
in each cluster.

%% file: 4-evalutations.tex
We next perform an empirical study of the effectiveness of the data drift measurement method proposed in \S\ref{sec:methodology}.
%In this section, we study the effectiveness of the data drift measurement method introduced in the previous section by performing empirical studies.
In particular, we use three real-world text datasets and 
apply our clustering-based drift measurement algorithm to 
quantify both synthetic and real drifts using several text embedding models. 

First we describe the datasets and embedding models used 
in our experiments, and the procedure used
for creating synthetic drift.
Then we present 
%the details of 
three sets of 
experiments in which we
(i) compare the sensitivity  of different embedding models to data drift,
(ii) study the effect of 
number of clusters on drift measurement, and 
(iii) study the effect of 
embedding dimensions on drift measurement.

In brief, across three datasets, we observe:
\begin{itemize}
\item The drift sensitivity of Word2Vec is generally poor. 
\item TF-IDF and BERT perform well in certain scenarios and badly in others.
\item The other embedding models -- Universal Sentence Encoder, Ada-001, and Ada-002 -- all have good performance, though the none of them was consistently the best.
\item Sensitivity improves (approximately) monotonically with the number of bins, $k$, but reaches a point of diminishing returns in the range $6\le k \le10$ across all datasets and models tested.
\item Increasing the size of embedding vector also improves model performance monotonically, beginning to saturate around 256 components.
\item (Interestingly) for models with large embedding sizes (e.g., Ada-002), drift saturation occurs at a much lower number of dimensions when sampled randomly.  This suggests that their sensitivity is not directly connected to their embedding size and that significant redundancy may be present.
\end{itemize}

\subsection{Datasets}
% \begin{figure*}[t]
% \centering
% \begin{tabular}{ccc}
% \includegraphics[width=0.33\textwidth]{figures/drift_sensitivity/20newsgroup_drift.png} &
% \includegraphics[width=0.33\textwidth]{figures/drift_sensitivity/civil_comments_drift.png} &
% \includegraphics[width=0.33\textwidth]{figures/drift_sensitivity/amazon_reviews_drift.png} \\
% (a) 20Newsgroup & (b) Civil Comments & (c) Amazon Reviews
% \end{tabular}
% \caption{Comparison of Different Embeddings}
% \label{fig:drift_sensitivity}
% \end{figure*}

Three real-world text datasets are used to perform the following experiments:\\

\noindent \textbf{20newsgroup} is available in the scikit-learn package. It has 18K news group posts and a target label indicating to which of 20 groups each post belongs. We only consider the training subset and group the data points into five general categories (``science'', ``computer'', ``religion'', ``forsale'', and ``recreation'') leaving 8966 news text and their corresponding general category label.\\

\noindent \textbf{Civil comments} \cite{beery2020iwildcam} is part of the WILDS\footnote{https://wilds.stanford.edu/datasets/} dataset.
% It has 447998 comments, and we sample out 10000 from those.
It contains online comments where a toxicity binary target and a demographic identity from 8 categories (``male'', ``female'', ``LGBTQ'', ``Christian'', ``Muslim'', ``other religions'', ``Black'', and ``White'') are assigned to every comment.\\
%During preprocessing, we put a binary threshold on the 8 categories.\\

\noindent \textbf{Amazon Fine Food Reviews} \cite{mcauley2013amateurs}
consists of reviews of fine foods from Amazon over a ten year period ending October 2012. Reviews include product and user information, ratings, and a plain text review\footnote{https://www.kaggle.com/datasets/snap/amazon-fine-food-reviews}.

\subsection{Text Embeddings}
We chose three popular embedding baselines namely Word2Vec \cite{mikolov2013distributed}, BERT \cite{devlin2018bert}, and Universal Sentence Encoder \cite{cer2018universal}. We also use embeddings generated by Large Language Models hosted by OpenAI under the model names, text-embedding-ada-001 and text-embedding-ada-002\footnote{https://platform.openai.com/docs/guides/embeddings} \cite{brown2020language}. Additionally, for comprehensiveness we also consider sparse representations generated by Term Frequency-Inverse Document Frequency (TF-IDF).

\subsection{Simulating Data Drift Measurement Scenarios}
We next explain how we use the three real-world datasets
to simulate different data drift measurement scenarios.
For two of the datasets, 20newsgroup and Civil comments, we
use the metadata information to 
%manufacture synthetic 
simulate drift that may be encountered during deployment. For the Amazon Reviews dataset, we use reviews from  2006 and 2007 as the baseline 
% against which we measure data drift in subsequent years.\\
and reviews from subsequent years as production data.

% The below approach will create a set of drift-induced datasets that can be used to test the robustness of a machine learning model to concept drift. By varying the percentage of drift labels/tags, the researcher can simulate different levels of concept drift.

\noindent \textbf{Generating synthetic data drift in 20newsgroup dataset.}
For this dataset, 
%we follow the following steps:
we adopt the following procedure:
\begin{itemize}
    \item We randomly split the dataset into a baseline set and a production set.
    % from which we sample data points that simulate deployment data. 
    The baseline set is 40\% of the total data points.
    \item  We chose a subset of categories and
    used them to introduce synthetic drift by either oversampling or undersampling these selected categories.
    % modifying the relative 
    % number of samples drawn from the selected category.
    % \item Separate the data points from the testing set that belongs to the selected group. The remaining data points are labeled as ``others''.
    % \item  We generate different scenarios of deployment data by modifying the percentage of samples that are drawn from the selected category versus the others category. Each production dataset contains 4000 data points in total.
    \item  We generate different scenarios of data drift by modifying the relative percentage of samples that are drawn from the selected categories. Each production scenario contains 4000 data points in total.
\end{itemize}

\noindent \textbf{Generating synthetic data drift in Civil comments
dataset.}
For this dataset, % we follow the following steps:
we adopt the following procedure:
\begin{itemize}
    \item We split the dataset into a baseline set and production set. The baseline contains 40\% of the total data points.
    % \item We chose the tag(s) based on which you want to manufacture synthetic drift.
    % We present results for distributional changes using the 
    % percentage of the ``female'' tag.
    \item We selected a subset of tags and used them to introduce synthetic drift by either oversampling or undersampling these selected tags.  
    % \item Separate the comments from the testing set for which the selected tag(s) is set to 1.
    % The remaining comments will be classified as ``others''.
    % \item For each drift scenario, we 600 data points are sampled from the selected tag(s) and the others category according to the specified percentages. 
    \item We generate different scenarios of data drift by modifying the relative percentage of samples that are drawn from the selected tags. Each production scenario contains 600 data points in total.
\end{itemize}

%%For the Amazon reviews dataset, choose the years of reviews to form the baseline. In this case, the years 2006 and 2007 were chosen. We randomly sample 1,500 reviews from the chosen years to form the baseline dataset. The remaining reviews from the chosen years from the production data.
% Since the Amazon reviews dataset is real-world yearly data; there is no need to induce drift. Instead, the researcher can observe whether their way of calculating drift would be able to catch the drift in the upcoming years, if there is any.

\subsection{Comparing Drift Sensitivity of Embedding Models}
\begin{figure*}[ht]
  \centering
  \begin{subfigure}{0.33\textwidth}
    \includegraphics[width=\linewidth]{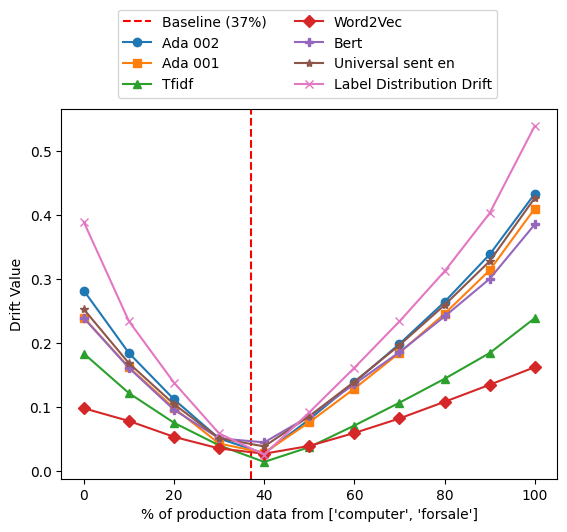}
    \caption{20Newsgroup}
    \label{fig:20newsgroup_drift}
  \end{subfigure}
  \hfill
  \begin{subfigure}{0.33\textwidth}
    \includegraphics[width=\linewidth]{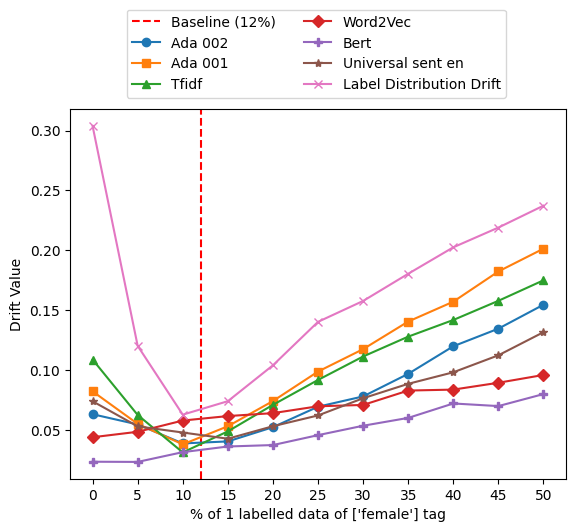}
    \caption{Civil Comments}
    \label{fig:civil_comments_drift}
  \end{subfigure}
  \hfill
  \begin{subfigure}{0.33\textwidth}
    \includegraphics[width=\linewidth]{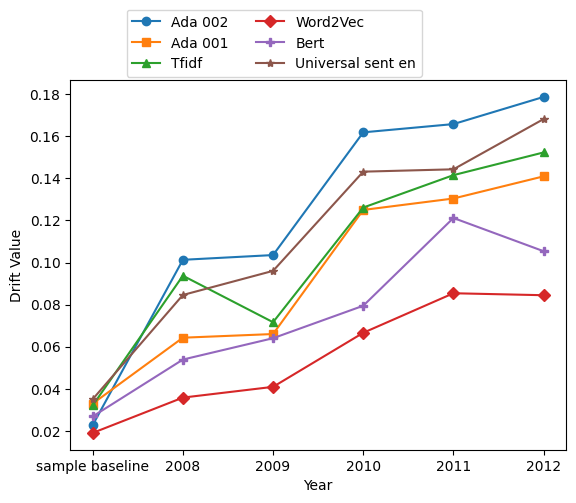}
    \caption{Amazon Reviews}
    \label{fig:amazon_reviews_drift}
  \end{subfigure}
  \caption{Drift sensitivity of embedding models.}
  \label{fig:drift_sensitivity}
\end{figure*}

In this experiment, we compare the sensitivity 
of our drift measurement algorithm when applied to
six different embeddings of different production data that correspond to drift scenarios introduced previously.
In order to control the effect of the embedding vector size, in this experiment we
use 300 dimensional vectors for all embeddings (larger embedding vectors are truncated at this length). Additionally, the number of clusters is set to 10.

Figure \ref{fig:drift_sensitivity} presents the results of this experiment for 
each of the three datasets.
In the synthetic data drift scenarios,
we also plot the drift value in the actual label distributions.
Furthermore,
% in the synthetic data drift scenarios,
the vertical line indicates the data distribution that 
corresponds to the distribution of the baseline data.

We find that Universal Sentence Encoder, Ada-001, and Ada-002 all perform quite well and that word2vec performs consistently poorly.  TF-IDF and BERT are inconsistent.  This could be related to specific vocabulary or model settings.

\subsection{Effect of Number of Clusters}
\begin{figure*}[ht]
  \centering
  \begin{subfigure}{0.33\textwidth}
    \includegraphics[width=\linewidth]{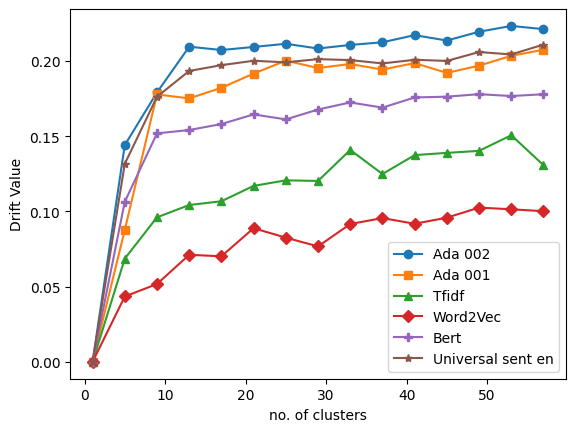}
    \caption{20Newsgroup : 60\% data from science}
    \label{fig:20newsgroup_cluster_saturation}
  \end{subfigure}
  \hfill
  \begin{subfigure}{0.33\textwidth}
    \includegraphics[width=\linewidth]{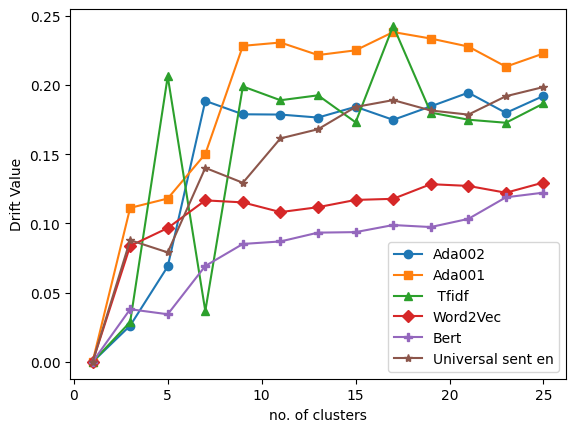}
    \caption{Civil Comments : 60\% of 1 labelled female}
    \label{fig:civil_comments_cluster_saturation}
  \end{subfigure}
  \hfill
  \begin{subfigure}{0.33\textwidth}
    \includegraphics[width=\linewidth]{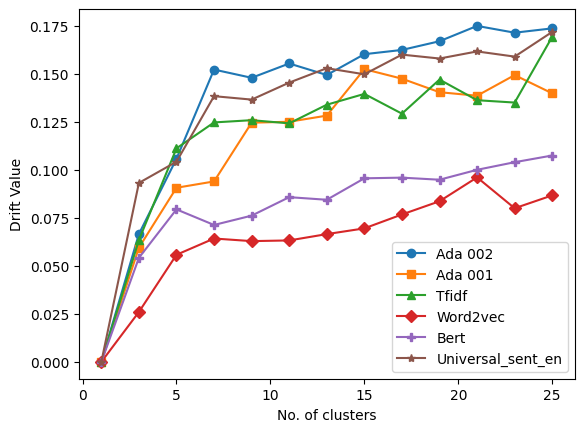}
    \caption{Amazon Reviews : Year 2010}
    \label{fig:amazon_reviews_cluster_saturation}
  \end{subfigure}
  \caption{Cluster saturation of different embeddings.}
  \label{fig:cluster_saturation}
\end{figure*}

% Is there any saturation, or we cannot see a significant increment in the drift value after a point due to increased clusters? To answer the above question, we are experimenting with the six different embeddings on the number of clusters for detecting the drift on the 3 datasets. 

While increasing the number of clusters generally increases the 
sensitivity to drift, we expect there to be saturation point after which
improvement diminishes rapidly.
 
For 20Newsgroup (Figure \ref{fig:20newsgroup_cluster_saturation}), the production dataset was comprised of a fixed 60\% from the ``science'' category and the remainder from others. We observe that from beyond 10 clusters, there was not a significant improvement in drift sensitivity. This observation informed our choice of 10 clusters for the sensitivity assessment in the prior section.

For Civil Comments (Figure \ref{fig:civil_comments_cluster_saturation}), the production dataset was comprised of 60\% of 600 examples with the female label and the remainder from the others. Here, also around 10 clusters, there was not any significant push in the drift detection.

For Amazon Reviews (Figure \ref{fig:amazon_reviews_cluster_saturation}), the production dataset was taken to be the 2010 reviews. Saturation occurs around 6-7 clusters.

In conclusion, we found that there is a saturation point for drift detection with increasing clusters. The number of clusters required for optimal drift detection depends somewhat on the dataset and the embeddings used but is in a narrow range around between $k=6$ and $10$.   We also observe instability in the TF-IDF plots.  This may be the result of specific tokens having special significance and falling across cluster boundaries in important ways.  In comparison, information of particular importance for the embedding models could be better distributed over their components.

% However, the LLMs could detect the most drift in all three datasets, regardless of the number of clusters.

\subsection{Effect of Embedding Dimension}

\begin{figure*}[ht]
  \centering
  \begin{subfigure}{0.33\textwidth}
    \includegraphics[width=\linewidth]{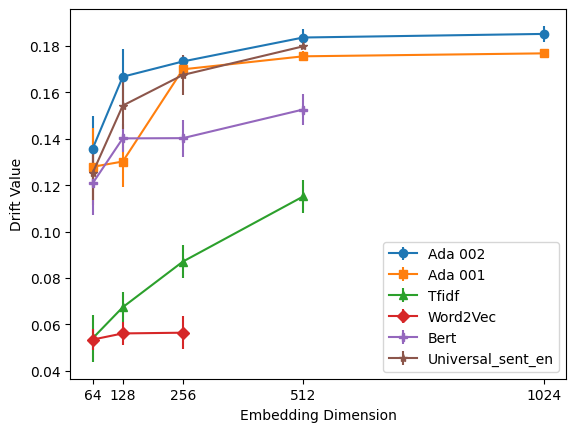}
    \caption{20Newsgroup: 60\% data from science}
    \label{fig:20newsgroup_embed_dim}
  \end{subfigure}
  \hfill
  \begin{subfigure}{0.33\textwidth}
    \includegraphics[width=\linewidth]{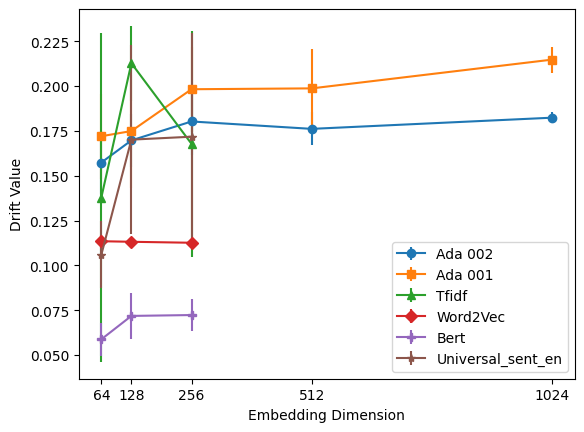}
    \caption{Civil Comments: 60\% of 1 labelled female}
    \label{fig:civil_comments_embed_dim}
  \end{subfigure}
  \hfill
  \begin{subfigure}{0.33\textwidth}
    \includegraphics[width=\linewidth]{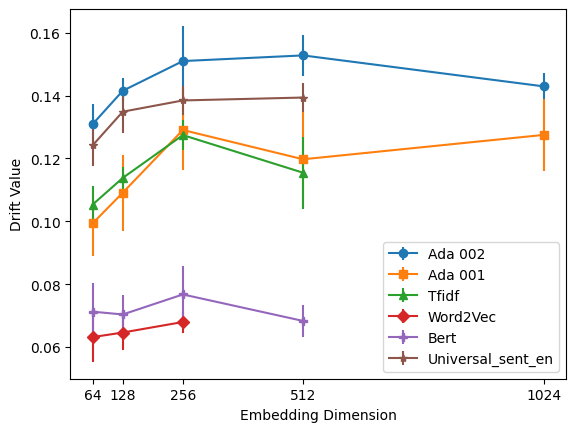}
    \caption{Amazon Reviews: Year 2010}
    \label{fig:amazon_reviews_embed_dim}
  \end{subfigure}
  \caption{Effect of embedding dimension on drift sensitivity.}
  \label{fig:dimension_exp}
\end{figure*}

Different models generate embedding vectors of different lengths.  Ada-002's embedding vectors have 1536 components, Ada-001's have 1024, BERT's have 768, and Universal Sentence Encoder's have 512. The length of TF-IDF vectors correspond to the number of tokens – we've chosen to take 300. To provide plots vs. embedding length, we randomly sample a set of $d$ components for each model whose whose embedding vectors  have a length $\geq d$.  In order to capture the uncertainty introduced by this sampling as well as variation in the $k$-means procedure, we perform the dimension sampling and drift calculation five times for each value of $d$, reporting the mean and standard deviation.  We define the ``production data'' as we do in the previous section and sweep over this embedding vector length $d$.

We find that the vector length, $d$, reaches a point of diminishing returns across the datasets and models around 256 components, regardless of the native size of the model's output.  This suggests that the sensitivity of the large models in not due to their large embedding size and that they may be storing redundant information.  In principle, this could also be connected to the fixed 10 cluster procedure.  We plan to explore this in future work.  As in the previous section we see see some instability in the TF-IDF scans.

%% file: 5-case-study.tex
The goal of this case study is to investigate a concrete scenario involving a real-world NLP model and demonstrate how data scientists can use our framework (deployed in the Fiddler ML Monitoring platform) to detect input data drift for a multi-class classification NLP model trained over the 20 Newsgroups dataset.
%We will demonstrate how data scientists can use Fiddler to monitor a real-world NLP model. 

\begin{figure*}[t]
  \centering
  \includegraphics[width=0.9\linewidth]{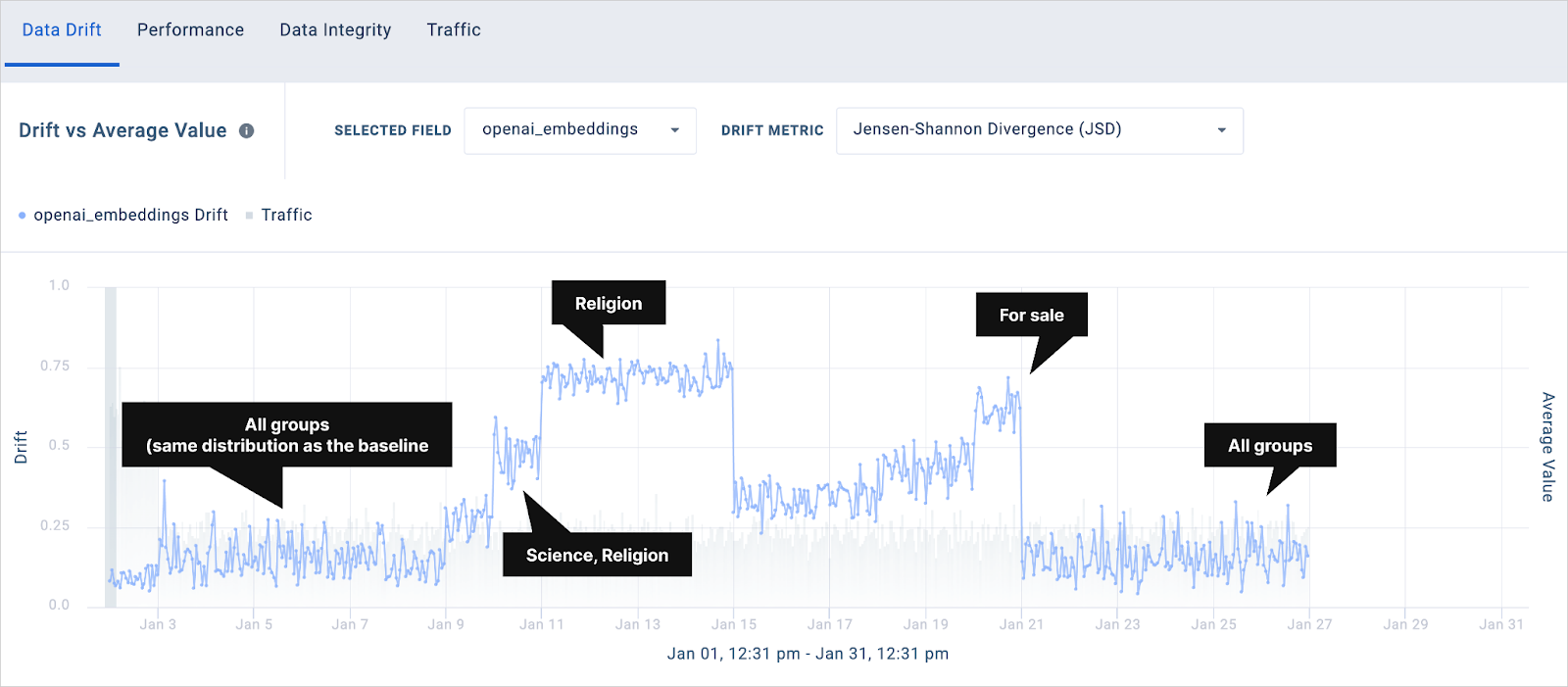}
  \caption{Monitoring OpenAI embeddings reveals distributional shifts in the samples drawn from the 20 Newsgroups dataset.}
  \label{fig:newsgroups-prodcut-view}
\end{figure*}

As noted earlier, the 20 Newsgroups dataset contains labeled text documents from different topics. We use OpenAI embeddings to vectorize text data, and then train a multi-class classifier model that predicts the probability of each label for a document at production. In particular, we obtain embeddings by querying OpenAI's text-embedding-ada-002 model (which is a hosted LLM that outperforms its previous models) using the Python API.

%First we need to vectorize the text data. OpenAI has recently published its latest text embedding model, text-embedding-ada-002, which is a hosted large language model (LLM) and outperforms its previous models. Furthermore, Open AI embedding endpoints can be easily queried via its Python API, which makes it an easy and efficient tool for organizations who want to solve NLP tasks quickly.

We keep the classification task simple by grouping the original targets into five general class labels: `computer', `for sale', `recreation', `religion', and `science'. Given the vectorized data and class labels we train a model using a training subset of the 20 Newsgroups dataset.

For monitoring purposes, we typically use a reference (or baseline) dataset with which to compare subsequent data. We create a baseline dataset by randomly sampling 2500 examples from the five subgroups specified in the 20 Newsgroup dataset.

To simulate a data drift monitoring scenario, we manufacture synthetic drift by adding samples of specific text categories at different time intervals in production. Then we assess the performance of the model in the Fiddler ML monitoring platform and track data drift at each of those time intervals. The implementation of our NLP monitoring framework (\S\ref{sec:methodology}) in the Fiddler platform supports easy integration of different NLP embedding model APIs (such as the ones provided by OpenAI). The user just needs to specify the input columns to be monitored (by defining a ``custom feature'' for NLP embeddings using the Fiddler API), and the associated tasks (obtaining the embeddings, performing clustering, and computing the data drift metric such as JSD) are handled by the platform.

%Now we present how Fiddler provides quantitative measures of data drift in text embeddings via Fiddler NLP (Embedding) Monitoring. This capability is designed to directly monitor the high-dimensional vector space of unstructured data. Therefore, NLP embedding models such as OpenAI can be easily integrated into Fiddler and users can start monitoring them without any additional work. All the user needs to do is to specify the input columns to a model that correspond to the embeddings vectors. This can be done by defining a ``custom feature'' for NLP embeddings using the Fiddler client API.

Figure \ref{fig:newsgroups-prodcut-view} shows the data drift chart within the Fiddler platform for the 20 Newsgroups multi-class model. More specifically, the chart is showing the drift value (in terms of JSD) for each interval of production events, where production data is modified to simulate data drift. The call outs show the list of label categories from which production data points are sampled in each time interval.

\subsection{Gaining more insights into data drift using UMAP visualization}

\begin{figure*}[t]
  \centering
  \begin{subfigure}{0.27\textwidth}
    \includegraphics[width=\linewidth]{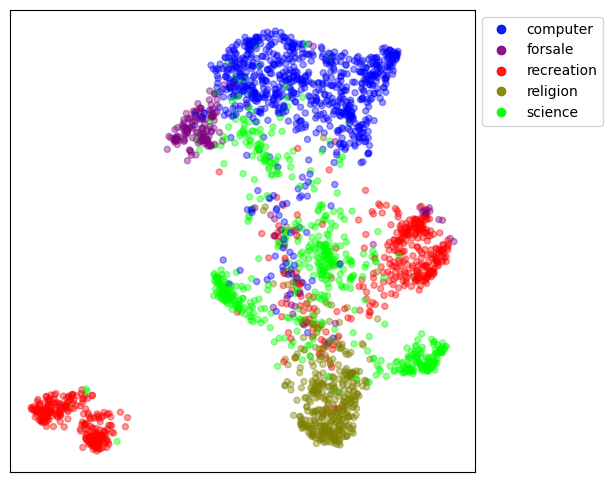}
    \caption{}
    \label{fig:umap1}
  \end{subfigure}
  \hfill
  \begin{subfigure}{0.34\textwidth}
    \includegraphics[width=\linewidth]{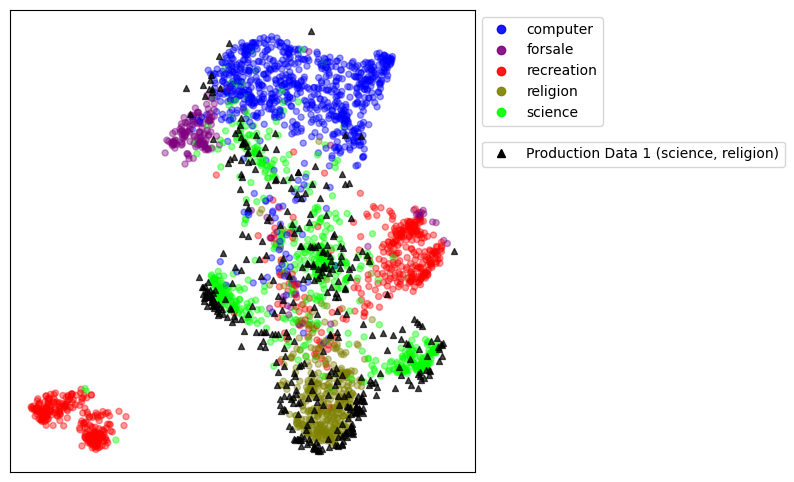}
    \caption{}
    \label{fig:umap2}
  \end{subfigure}
  \hfill
  \begin{subfigure}{0.32\textwidth}
    \includegraphics[width=\linewidth]{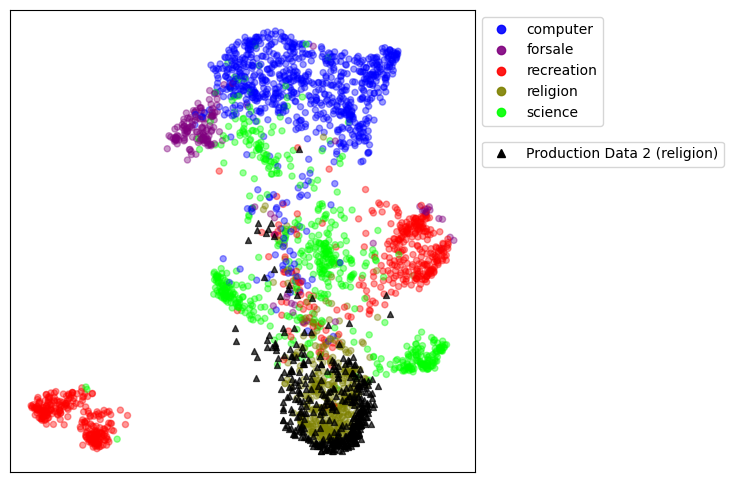}
    \caption{}
    \label{fig:umap3}
  \end{subfigure}
  \caption{2D UMAP embeddings of baseline and production vectors obtained from OpenAI}
  \label{fig:umap}
\end{figure*}

In the monitoring example presented above, since data drift was simulated by sampling from specific class labels, we could recognize the intervals of large JSD value and associate them with known intervals of manufactured drift. However, in reality, oftentimes the underlying process that caused data drift is unknown. In such scenarios, the drift chart is the first signal that is available about a drift incident which can potentially impact model performance. Therefore, providing more insight about how data drift has happened is an important next step for root cause analysis and maintenance of NLP models in production.

The high-dimensionality of OpenAI embeddings (the Ada-002 embeddings have 1536 dimensions) makes it challenging to visualize and provide intuitive insight into monitoring metrics such as data drift. In order to address this challenge, we use Uniform Manifold Approximation and Projection (UMAP) \cite{McInnes2018} to project OpenAI embeddings into a 2-dimensional space while preserving the neighbor relationships of the data as much as possible.

Figure \ref{fig:umap1} shows the 2D UMAP visualization of the baseline data colored by class labels. We see that the data points with the same class labels are well-clustered by UMAP in the embedded space although a few data points from each class label are mapped to areas of the embedded space that are outside the visually recognizable clusters for that class. This is likely due to the approximation involved in mapping 1536-dimensional data points into a 2D space.  Another explanation is that the Ada-002 embedding model has identified semantically distinct subgroups within topics.

In order to show how UMAP embeddings can be used to provide insight about data drift in production, we will take a deeper look at the production interval that corresponds to samples from ``science'' and ``religion'' categories. Figure \ref{fig:umap2} shows the UMAP projection of these samples into the UMAP embeddings space that was created using the baseline samples. We see that the embedding of unseen data is aligned fairly well with the regions that correspond to those two class labels in the baseline, and a drift in the data distribution is visible when comparing the production data points and the whole cloud of baseline data. That is, data points are shifted to the regions of space that correspond to ``science'' and ``religion'' class labels.

Next, we perform the same analysis for the interval that contains samples from the ``religion'' category only, which showed the highest level of JSD in the drift chart in Figure \ref{fig:newsgroups-prodcut-view}. Figure \ref{fig:umap3} shows how these production data points are mapped into the UMAP space; indicating a much higher drift scenario.

Notice that although UMAP provides an intuitive way to track, visualize and diagnose data drift in high-dimensional data like text embeddings, it does not provide a quantitative way to measure a drift value. On the other hand, our clustering-based NLP monitoring approach provides data scientists with a quantitative metric for measuring drift, and thereby enables them to configure alerts to be triggered when the drift value exceeds a desired threshold.

%% file: 6-use-cases.tex
We next present the key insights and lessons learned from deploying the framework described in the previous sections as part of 
the Fiddler ML monitoring platform \cite{fiddler_monitor} over a period of 18 months.
%%a model monitoring platform over a period of more than one year. 
%The system detailed in the previous sections was deployed to assess the utility for real-world use-cases. 
In one such real-world deployment, the system was tasked with detecting distributional shifts in the text queries being typed by the users into the search bar of a popular 
online design platform.
%website. 
The text queries were encoded into a 384-dimensional embedding vector with the help of a deep neural network (trained on the proprietary data of the design platform). The cluster centroids and the baseline distribution were determined on a baseline dataset of 500K queries. The production data consisted of 1.3M queries over a period of 2.5 months. Drift was then computed with a window size of 24 hours and plotted as a time-series (corresponding to one JSD value for each day of the 2.5 month period). Using this approach, the monitoring system flagged a bump in drift that was first detected in the fifth week of traffic and persisted from then on. Inspecting the cluster-based histograms pointed to one cluster which contributed to most of the observed drift. It turned out that this particular cluster corresponded to queries that were either gibberish or from a language other than English. On further investigation, it was revealed that certain data pipeline changes were made for such queries during the fifth week, causing the new production distribution to differ significantly from the baseline distribution.

During the course of development and deployment of the system, we realized that users benefit from not just knowing the JSD values 
%for different production time intervals 
and getting alerts when these values exceed the desired thresholds, but also from the ability to analyze the underlying reasons for the drift. In particular, users expressed interest in associating semantically meaningful information for each cluster. We addressed this need by (1) highlighting representative examples from each cluster
%(which could be viewed by clicking or hovering over each cluster in the product visualization)
and (2) providing a human interpretable summary of the cluster.
We decided to provide such a summary in the form of the top distinctive terms (obtained using TF-IDF applied over the set of text data points in the cluster) since the resulting summary was informative and could be generated with low cost and latency.
%We found that this option was preferable to a natural language summary generated by a cloud-based LLM (due to cost / privacy considerations).
Further, we observed that providing a visualization tool that plots baseline and production samples using UMAP \cite{McInnes2018}, a dimensionality reduction technique, enables the user to interactively perform debugging. The key insight is that such analysis, debugging, and visualization tools, along with the technical approach outlined in this paper, can help the user to fully benefit from the system -- namely, to not just observe drift values over time and get alerts but also diagnose and mitigate the underlying causes.

%% file: 8-conclusion.tex
Motivated by the increasing adoption of NLP models in practical applications and the need for ensuring that these models perform well and behave in a trustworthy manner after deployment, we proposed a clustering-based framework to measure distributional shifts in natural language data and showed the benefit of using LLM-based embeddings for data drift monitoring. We introduced sensitivity to drift as a new evaluation metric for comparing embedding models and demonstrated the efficacy of our approach over three real-world datasets. We implemented our system as part of the Fiddler ML Monitoring platform, and presented a case study, insights, and lessons learned from deployment over a period of 18 months. As our approach has been developed, evaluated, and deployed to address the needs of customers across different industries, the insights and experience from our work are likely to be useful for researchers and practitioners working on NLP models as well as on LLMs.

Our findings shed light on a novel application of LLMs -- as a promising approach for drift detection, especially for high-dimensional data, and open up several avenues for future work. A few promising directions include: (1) whether domain-specific LLM embeddings are desirable for detecting drift in a given NLP application, (2) combining information across multiple NLP models in an application setting to detect drift, and (3) exploring embedding based drift detection approaches for image, voice, and other modalities as well as multi-modal settings.